\def\year{2017}\relax
\documentclass[letterpaper]{article}
\usepackage{aaai17}

\usepackage{times}
\usepackage{url}
\usepackage{latexsym}
\usepackage{float,color}
\usepackage{epsfig}
\usepackage{epstopdf}
\usepackage{url}
\usepackage{hyperref}
\usepackage{graphicx,subfigure}
\usepackage{threeparttable}
\usepackage{booktabs}
\usepackage{enumitem}
\usepackage{amsmath}
\usepackage{amsfonts}

\newcommand{\ignore}[1]{}
\usepackage{todonotes}

\begin{document}
	%
	\title{Incidental Supervision:\\ Moving beyond Supervised Learning}
	\author{Dan Roth\\
		Department of Computer Science \\
        University of Illinois at Urbana/Champaign \\
		\tt{danr@illinois.edu} \\
	}
\maketitle

\begin{abstract}

Machine Learning and Inference methods have become ubiquitous in our attempt to induce more abstract representations of natural language text, visual scenes, and other messy, naturally occurring data, and support decisions that depend on it. However, learning models for these tasks is difficult partly because generating the necessary supervision signals for it is costly and does not scale.

This paper describes several learning paradigms that are designed to alleviate the supervision bottleneck. It will illustrate their benefit in the context of multiple problems, all pertaining to inducing various levels of semantic representations from text.
In particular, we discuss  (i)
{\em Response Driven Learning} of models, a learning protocol that supports inducing meaning representations simply by observing the model's behavior in its environment, (ii) the exploitation of {\em Incidental Supervision} signals that exist in the data, independently of the task at hand, to learn models that identify and classify semantic predicates, and (iii) the use of weak supervision to combine simple models to support global decisions where joint supervision is not available.

While these ideas are applicable in a range of Machine Learning driven fields, we will demonstrate it in the context of several natural language applications, from (cross-lingual) text classification, to Wikification, to semantic parsing.

\end{abstract}

\section{Introduction} \label{sec:intro}

The fundamental issue underlying natural language understanding is that of Semantics.
There is a need to move toward understanding the text at an appropriate level of abstraction, beyond the word level, in order to support access, knowledge extraction and communication.

In all these cases there is a need to get around the inherent {\em ambiguity} in natural language expressions -- every utterance might carry multiple meanings, depending on the context in which it is being used, the audience, the producer of the utterance, etc. -- and the {\em variability} in natural language -- a desired meaning can be expressed in a very large number of quite different surface forms.

This necessitates the use of machine learning and inference methods to support inducing the desired level of abstraction.  Inducing models and semantic representations and making decisions that depend on these require learning and, in turn, supervision.
Given a task, the standard machine learning methodology suggests to collect annotated data {\em for this task}, and then train a model for it.
However, we believe that this methodology is not scalable -- we will never have enough annotated data to train all the models we need this way. Annotating data for complex tasks is difficult, costly, and sometimes impossible, especially when a required intermediate representation is ill-defined, but the outcome building on it is.
The popular direction of using crowd-sourcing, while often important and helpful in relatively small tasks, is not a realistic solution when an annotation of a single instance takes an expert 5 minutes -- a common situation when dealing with annotating involved meaning representations for semantic parsing, or relations between events for an information extraction task.

This paper suggests to re-think the current annotation-heavy approaches to Natural Language Processing and many other areas that make use of machine learning methods to deal with messy, naturally occurring data. Specifically, we argue that in many realistic cases learning should be (and is) driven by {\em incidental signals}. Incidental Signals refer to a collection of weak signals that {\em exist} in the data and the environment, independently of the tasks at hand. These signals are co-related to the target tasks, and can be exploited, along with appropriate algorithmic support, to provide sufficient supervision and facilitate learning.

\begin{figure}
\centering
\includegraphics[width=225px]{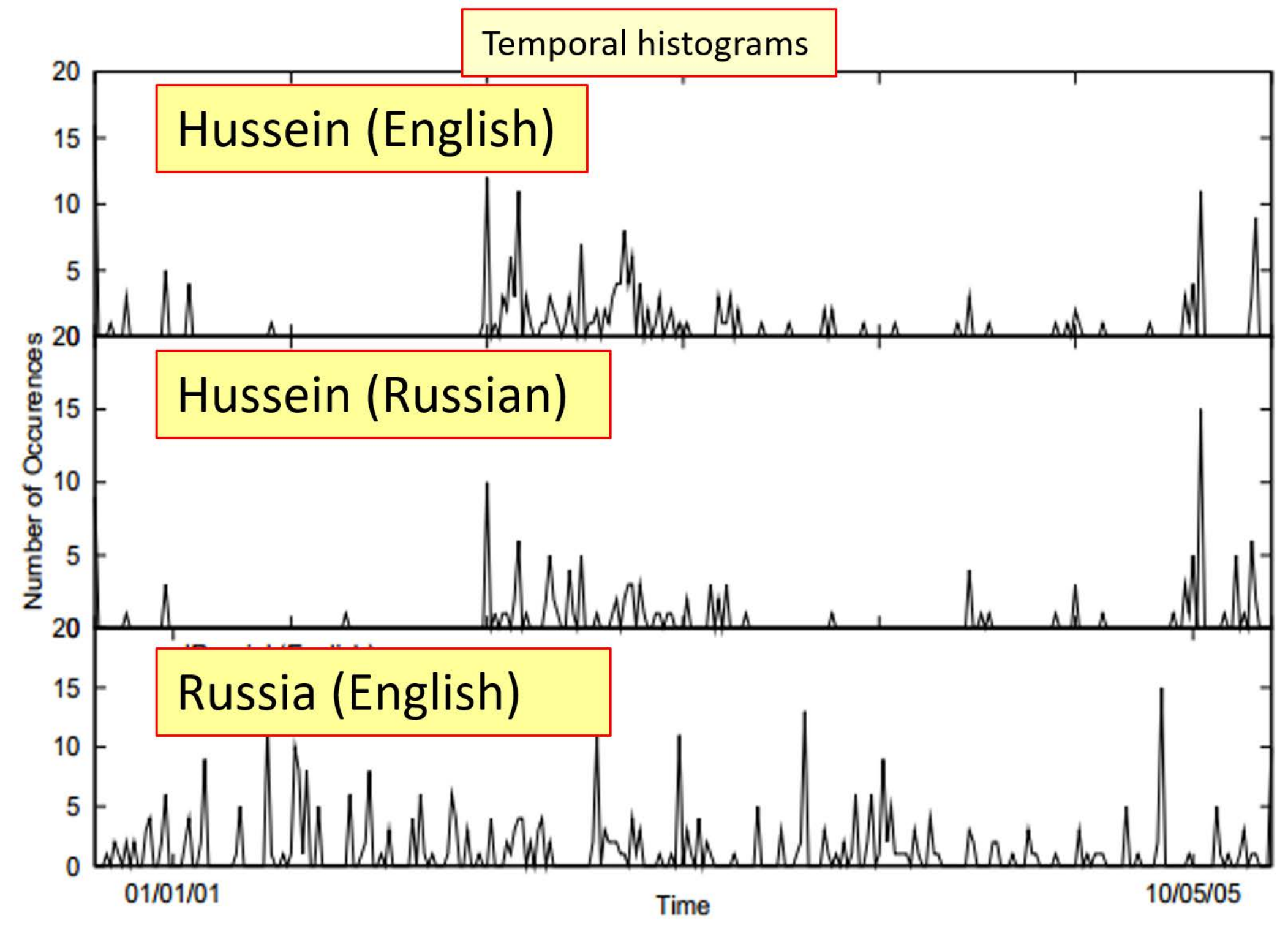}
\caption{Temporal Alignment of NEs. A strong signal (Hussein (E); Hussein (R)) and no signal (Hussein, Russia).}
\label{fig:trans}
\end{figure}

Consider, for example, the task of Named entity (NE) transliteration  -- the process of transcribing a NE from a source language to some target language based on phonetic similarity between the entities (e.g., determine how to write "Obama" in Hebrew). Identifying transliteration pairs is an important component in many linguistic applications which require identifying out-of-vocabulary words, such as machine translation and multilingual information retrieval. Naturally, to know how to write the word {\em Hussein} in Russian, say, one needs to train a model that requires as input a long list of pairs: NEs in English and their corresponding transliteration in Russian. These resources do not exists for many pairs of languages. However, \cite{KlementievRo06} showed
that it is possible to automatically discover NEs in low resource languages, given bilingual corpora that are weakly temporally aligned. This is a much easier resource to come by, given that multiple news services generate comparable news data in many languages. Moreover, this resource is independent of the task at hand and can be used to aid supervising multiple related tasks. Mentions of NEs that correspond to the same entity, in different languages, would then have similar temporal distributions across such corpora, and this similarity is a strong signal that a pair of NEs could be a transliteration of each other.
Figure~\ref{fig:trans} illustrates this situation by showing the temporal histogram of the NE "Hussein" over the same time period in a comparable English-Russian corpus, along with the lack of a signal when considering a  different NE.

Clearly, this signal by itself is not sufficient to supervise the learning of a transliteration model since the temporal distribution of {\em Hussein} might be quite similar to that of {\em Iraq} too. However, along with other weak signals such as phonetic similarity, it would support learning a good model.

This example provides an illustration of what we mean by an {\em incidental supervision signal} in this paper. The temporal signal is there, independently of the transliteration task at hand. It is co-related to the task at hand and, together with other signals and some inference, could be used to supervise it without the need for any significant annotation effort.

We note that our notion of supervision is different from that of {\em distant supervision}~\cite{MBSJ09}
where a model is learned given a labeled training set, as in ``standard'' supervised machine learning, but the training data is labeled automatically based on heuristics. In the example above, a complete training set never exists, and the algorithm needs to make use of multiple weak signals.

The rest of the paper will provide examples of incidental supervision from several lexical and structural tasks in natural language semantics, and argue for the need and the possibility of an incidental supervision framework by describing several dimensions of it.
Specifically, we will present three types of incidental supervision scenarios:
(i) Exploiting incidental cues in the data, unrelated to the task, as sources of supervision; (ii) Learning complex models in the absence of complete annotation, by reasoning over the outcomes of simpler, easier to learn, models (with the aid of some declarative knowledge); and (iii) Supervising models indirectly, by providing feedback based on the behavior of the model in the world.

\section{Understanding the Label Space} \label{sec:dataless}

In this section we discuss several text classification tasks, at varying levels of complexity. In each case, we illustrate how to induce models for these without directly training on the target task. Instead, we show that a good understanding of the {\em label space} can be acquired independently of the task at hand, and that this understanding can be used to develop high quality learning models.

The simplest example of this can be shown in the task of text categorization.
Traditionally, text categorization has been studied as a problem of training a multiclass classifier using labeled data~\cite{Sebastiani02}. However, humans can categorize documents into named categories without any explicit training, relying on their understanding of the meaning of category names. However, with the vast amount of information on the web it is possible today, given an taxonomy of labels (possibly coupled with label definitions that disambiguate it if needed), to generate a good semantic representation for each label in a given taxonomy.

A series of papers by multiple authors~\cite{cRRS08,SongRo15,CYPC15} has developed this idea within a framework called {\em Dataless Classification}.
Given a single document along with a taxonomy of labeled categories into which one wants to classify the document, the dataless procedure proceeds as follows:
\begin{itemize}
\item Let $\Phi(l_i)$  be the semantic representation of label $l_i$.
\item Let $\Phi(d)$ be the semantic representation of document $d$.
\item Select the most appropriate category label according to:
$$ l_i^{*} = argmin_i d(\Phi(l_i), \Phi(d)),$$
\end{itemize}
where $d( \cdot )$ denotes an appropriate metric in the semantic space.
It is important to note that this is {\em not} an unsupervised learning scenario. Unsupervised learning assumes a coherent collection of data points, and that similar labels are assigned to similar data points~\cite{ZGBD09}. It {\em cannot work on a single document} as the aforementioned dataless process suggests. However, if the dataless method is given, instead, a {\em collection} of documents that is assumed to be coherent, it is possible to follow the initial classification step above with the standard semi-supervised learning procedure and bootstrap from the basic dataless classifier.
The dataless process is similar, but not identical, to related proposals that were developed at about the same time and were used mostly in the computer vision community, namely zero-shot and one-shot learning.
In the former (e.g.~\cite{PPHM09}), the goal is to learn a multiclass classifier with a range $Y=\{1,\ldots , k\}$ where only some of the labels are given in the training set. In the latter~\cite{FeiFeiFePe06}, one gets only a few training examples per-category and is expected to learn a multiclass classifier.
Dataless classification is similar to one-shot, if one wants to view the labels' definitions as a small number of examples, and to zero-shot since no training data is given, but in dataless the  target labels could be independent.

The key question underlying the success of dataless classification and the other related methods is: ``How can one generate good semantic representations?"
While there has been a lot of renewed interest over the last few years in semantic representations it turns out that, in the context of topical classification,  the most successful representation (e.g., \cite{SongR14}) is the {\em Extended Semantic Analysis (ESA)} representation, a Wikipedia based representation developed in ~\cite{GabrilovichMa09}. In this representation, each word is represented as a weighted (sparse) vector of Wikipedia titles that mention it.

The resulting learning approach thus requires no direct supervision for the task. It relies on the modeler to make use of incidental signals that exist independently of the learning task; in this case, the existence of Wikipedia, and the fact that the accumulation of Wikipedia pages a word appears in provides a good understanding of the meaning of this word.

This approach has been quite successful in supporting (hierarchical) topic classification. Moreover, other incidental supervision signals can be added to augment this functionality. For example, the existence of cross-lingual links between similar Wikipedia titles -- a Wikipedia page titled ``Basketball'' has a link to a corresponding Italian page ``Pallacanestro'', and another one to a Spanish page ``Baloncesto'', allows one to augment the dataless classification approach and classify documents in multiple languages into an English taxonomy of categories.

Many other models for textual classification tasks can be induced in a conceptually similar way. One key example is the task of Wikification.

The literature on Wikification, the task of identifying and grounding mentions of entities and concepts into encyclopedic resources, builds on the fact that the content of Wikipedia pages ``explains'' their title. Moreover, it counts on authors and editors to hyperlink (a subset of the) mentions in Wikipedia to the appropriate titles. This incidental supervision signal provides the ability to train ranking models that map a mention in its context to the appropriate Wikipedia title without any task specific annotation~\cite{RRDA11,Cucerzan07,MilneWi08,MihalceaCs07,ChengRo13}. Thus, a mention of {\em Clinton} in a specific context can be disambiguated to whether it is {\em Hillary Clinton, Bill Clinton, the Clinton Nuclear Generating Station} or any of the many other senses of {\em Clinton}, without any direct supervision. As pointed out above, with the additional incidental supervision signal provided by the cross-lingual links, documents in multiple languages can be Wikified into the English Wikipedia~\cite{TsaiRo16b}. %
Incidental supervision might  be even more important when trying to ground concepts into knowledge bases that cover multiple aspects of a domain -- as is the case in the medical domain~\cite{JimenoAr10}.
This problem poses a few additional challenges beyond those addressed in the popular Wikification setting mentioned above. Key among them is that most knowledge bases do not contain the rich textual and hyperlink information Wikipedia does; consequently, the main supervision signal used to train Wikification rankers does not exist. However, other incidental supervision signals exist and ~\citeauthor{TsaiRo16} (\citeyear{TsaiRo16}) show that the fact that a small percentage of the textual concepts have entries in multiple knowledge bases, redundant entries or just related entries, provides a sufficiently strong supervision signal to train a good ranking model. This is an illustration of a somewhat more sophisticated use of an incidental supervision signal, nevertheless it exhibits the notion that signals are out there, and they can be used to train a variety of models.

Other tasks in natural language processing, such as context sensitive spelling and grammar checking, have relied on incidental supervision for many years. Specifically, under the assumption that most edited textual resources (e.g., the New York Times, Wikipedia) have been carefully edited for spelling and grammar, these methods generate contextual representations for words, punctuation marks, and phenomena such as agreements, and then use these representations to identify mistakes and correct them in a context sensitive manner~\cite{RozovskayaRo14,GoldingRo99}.

These are, in fact, the same incidental supervision assumptions that underlie methods like LSI~\cite{DDFL90} and the recently popularized word embeddings~\cite{MikolovYiZw13}, with the key difference that in the former, a more accurate notion of negative examples is used to generate representations that, consequently, can be used directly as classifiers.

Finally we note that, in most cases, making use of multiple incidental supervision signals requires some inference mechanism that supports making decisions by putting the resulting models together~\cite{KlementievRoSm08,ChengRo13}.

\section{Structured Label Spaces} \label{sec:joint}

In this section, we discuss the problem of supervising learning models for structured output spaces. The challenge in learning structured models is that the prediction (also called decoding, or inference) involves assigning values to multiple interdependent variables where the expressive dependency structure -- modeled as a graph, a tree, or a sequence -- can influence, or even dictate, what assignments are possible.

We will use the example of (Extended) Semantic Role Labeling~\cite{PalmerGiKi05} to illustrate the key machine learning challenges. While the original semantic role labeling task was defined for semantic relations expressed by verb predicates, it has been clear that sentences express relations via other linguistic phenomena as well~\cite{SrikumarRo11,ArivazhaganChRo16}. Furthermore, these phenomena interact with each
other, thus restricting the structures they articulate.
The key challenge from the Machine Learning perspective is that large jointly labeled corpora do not exist and are extremely difficult to generate. However, for many of the individual phenomena of interest, including verb predicates, prepositional predicates, temporal relations, light verbs, compound nouns, etc., there exist independent collections of annotated text (largely since the independent annotation task is significantly easier).
The scarcity of jointly labeled data presents a crucial technical challenge
for learning a joint model and, as importantly, to scaling it up when additional phenomena are to be added, at later times.

The incidental supervision perspective of this problem would consider a model for each phenomena (or, potentially, tightly coupled phenomena for which jointly labeled data exists) as black boxes. It would then make use of coherency constraints between the predictions of these models as a way to support coherent global predictions that satisfy the interdependencies among the phenomena.

This can be done with minimal jointly labeled data to guarantee scaling of different models~\cite{SrikumarRo13}, but can also be done with no jointly labeled data, in a multi-view learning paradigm~\cite{GGBT07,Sun13}, where multiple output views of the same data can be learned together, with one model per view. There are multiple computational frameworks that support this type of processing. One way could be to consider the existing ``simple'' models as fixed components in a global constrained optimization problem, where each of these produces a probability distribution over its possible outputs. An alternative could be to propagate some feedback from the global solution back to individual components. This last approach can be viewed as an instance of Constraint Driven Learning~\cite{ChangRaRo07,ChangRaRo12,CRRR08} and its continuous counterpart, posterior regularization~\cite{GGGT10}, where multiple simple (distributional) signals contribute to a global decision, by being pushed to satisfy expectations on the global decision.

The perspective described in this section provides another important view on supporting involved, global decisions, via rather simple, indirect, supervision signals. This view depends on the decision making process being aware of the interdependencies among the expected predictions of their components but does not necessitate significant annotation effort for large structures.

\section{Response Driven Learning} \label{sec:response}


In this section we address a third perspective of indirect supervision, which we motivate via a rather involved natural language understanding task.

The reason we want to develop intelligent agents with natural language understanding abilities is to allow humans to communicate with the agent for the purpose of providing it, for example, some relevant domain expertise. Key requirement of this interaction is that we want to communicate with the intelligent agent without knowing anything about the internal representations used by the agent.

This natural language interpretation task, often referred to as {\em semantic parsing}, is typically formulated as that of learning a mapping between natural language input and a formal meaning
representation; an ``executable" representation that the agent can act on -- the agent could be a data base responding to a natural language query, a game API responding to natural language instructions, etc.
Technically, this problem can be formulated in a straight forward way as the following machine learning problem: given a set of input sentences and their corresponding meaning representations, learn a mapping -- a semantic parser -- that can map a new, previously unseen sentence into its meaning representation~\cite{ZelleMo96,ZettlemoyerCo05}.
However, this learning algorithm would require large amounts of annotated data to account for the expected variability in the input, and this annotation is costly, since it requires expertise in forming the logical meaning representations the agent can take. Moreover, this annotation process may have to be repeated when the interaction is done with a different agent that might use a different representation.

The incidental supervision perspective in this case is based on the observation that the target meaning representation is to be executed by a computer program -- the learning agent -- which in turn
provides a response or outcome. That is, there is a simple derivative of the meaning representation -- the action produced by the agent given this meaning representation. Consequently, the incidental supervision perspective suggests to use the fact that this derivative is bound to be naturally supervised in the environment, and exploit these indirect signals in the interaction between the learning agent and the environment (or teacher) rather than annotate meaning representations. This brings up the technical question of whether one can rely on this weak level of interaction to provide sufficient supervision and, eventually, support the recovery of the meaning representation.

More specifically, if we assume that the goal is to interact with a game API or a database, the simple derivative is the execution of the meaning representation on the API or the database. Consider transforming the input sentence into the executable meaning representation using the current model; once this representation is executed -- an instruction is being sent to the API in its own formal language, or a query is being issued to the database in its formal language -- feedback is provided at the level of: ``this API instruction is legitimate/good/bad'' or ``the answer supplied by the database is not the expected one'' (indicating that the semantic parsing process failed to produce the correct query). In a more abstract way, consider some simple derivatives of the semantic parser model's outputs; supervise the derivative (e.g., ``is the answer provided by the database correct?") and propagate it to learn the complex, structured, semantic parsing model.

Indeed, \cite{CGCR10} proposed a new learning paradigm -- {\em Response Driven Learning} --  capable of exploiting this level of incidental supervision, based on the easy-to-supervise response. The feedback can be viewed as a teacher judging whether the execution of the meaning representation produced the desired response for the input sentence.

We note that this learning paradigm is conceptually related, but technically different, from that of reinforcement learning and is also different from a simple end-to-end task done with a neural network. The goal here is to learn a complex structured model such as a semantic parser. And, it is important that this model, which is viewed as latent in the response driven learning paradigm, is executable, so that it yields a response from the environment or the teacher.

This type of supervision is natural in many
situations and requires no expertise, thus can be
supplied by any user. Indeed, more work is being done in this paradigm~\cite{LiangJoKl11,ArtziZe13} and the results show, perhaps surprisingly, that without using any annotated meaning representations,
learning with this weak incidental supervision signal is capable of producing a parser that is competitive with fully supervised parsers.

Technically, this mode of supervision, that relies on feedback from the behavior of the model in its environment, is more challenging than earlier incidental supervision instances we described. However, given that we typically want our models to be used, this supervision is bound to exist and be available, if we can address the challenge of using it well. In addition, it is important to note that the supervision protocol is independent of the learning model itself; thus, recent attempts to learn semantic parsers using sequence-to-sequence neural network models~\cite{DongLa16} could also be embedded within this more general and realistic training paradigm.

\section{Conclusion} \label{sec:conc}


This paper proposes that the AI community needs to re-think the current annotation-heavy approaches used in Natural Language Processing, Computer Vision, and many other areas that make use of machine learning methods to deal with messy, naturally occurring data. We argue that in order to scale up our use of learning methods and move forward in our ability to support complex cognitive computations, learning should be (and is) driven by {\em incidental signals}. These are weak signals that {\em exist} in the data and the environment, independently of the tasks at hand, and are co-related to the target tasks. We should develop ways to identify and exploit these signals and, along with it, develop the necessary inference support  so that we can provide sufficient supervision and facilitate learning without the need to annotate data for each task of interest.
We provided evidence in the form of three incidental supervision scenarios in a range of natural language understanding tasks, and gave a preliminary characterization of the type of incidental supervision scenarios the community might want to study further. Our research community needs to address these challenges by developing principled ways to identify these signals and by studying principled algorithmic approaches for using them.

\section*{Acknowledgments}
The authors would like the reviewers for comments that helped to improve this presentation.
This material is based on research sponsored by the US Defense Advanced Research Projects Agency (DARPA) under agreements FA8750-13-2-000 and HR0011-15-2-0025. The U.S. Government is authorized to reproduce and distribute reprints for Governmental purposes notwithstanding any copyright notation thereon. The views and conclusions contained herein are those of the authors and should not be interpreted as necessarily representing the official policies or endorsements, either expressed or implied, of DARPA or the U.S. Government.

{
\bibliographystyle{aaai}
\bibliography{newref,ccg,cited}
}

\end{document}